\pdfoutput=1
\documentclass[11pt]{article}

\usepackage{ACL2023}

\usepackage{times}
\usepackage{latexsym}

\usepackage[T1]{fontenc}
\usepackage[utf8]{inputenc}

\usepackage{microtype}

\usepackage{inconsolata}

\usepackage{gb4e}
\noautomath
\usepackage{multirow}
\usepackage{tikz}
\usepackage{adjustbox}
\usepackage{booktabs}

\title{SIGMORPHON 2023 Shared Task of Interlinear Glossing: Baseline Model}

\author{Michael Ginn \\
  University of Colorado \\
  \texttt{michael.ginn@colorado.edu}\\}

\begin{document}
\maketitle
\begin{abstract}
Language documentation is a critical aspect of language preservation, often including the creation of Interlinear Glossed Text (IGT). Creating IGT is time-consuming and tedious, and automating the process can save valuable annotator effort. 

This paper describes the baseline system for the SIGMORPHON 2023 Shared Task of Interlinear Glossing.\footnote{\url{https://github.com/sigmorphon/2023glossingST/tree/main/baseline}} In our system, we utilize a transformer architecture and treat gloss generation as a sequence labelling task. 
\end{abstract}
\section{Introduction}

Language documentation is a vital goal in linguistics, with over half of the world's languages classified as threatened or worse \citep{seifart_language_2018}. Documentation facilitates language teaching, analysis, and development of language technology. However, documentation is time-consuming and resource-expensive, particularly for endangered languages where the number of speakers who are available and willing to help document is few. Furthermore, a great portion of documentation is repetitive and monotonous for documenters, resulting in errors and omissions.

Thus, it is desirable to develop methods that automate as much of the work as possible, so that documentation can be done more quickly, for larger amounts of data, and for a wider number of languages. It is unlikely that any automated system will be able to produce documentation with near-perfect accuracy for low-resource languages; however, even a system with lower performance is beneficial when used in conjunction with a human annotator \citep{palmer2009evaluating}.

Transcribed text is produced from recorded spoken language, and generally is recorded either in the language's native orthography, a universal system such as the International Phonetic Alphabet (IPA), or an approximation in another language's orthography. Transcription for low-resource languages is often nonstandard and varies from transcriber to transcriber, resulting in noiser data.

Morphological segmentation refers to the process of dividing the recorded text into morpheme units, which linguists define to be the smallest meaning-bearing unit of language. For example, the English word \textit{cats} would be segmented into \mbox{\textit{cat-s}}, where \textit{-s} is the plural morpheme. Segmentation is particularly valuable for agglutinative and poly-synthetic languages, where words may be formed of many morphemes.

\subsection{Interlinear Glossed Text}

Linguistic annotation refers to the process of producing a format such as Interlinear Glossed Text, which records the syntactic and morphological properties of words in a corpora. IGT is not standardized and varies from annotator to annotator, depending on the linguistic information they wished to convey \citep{palmer2009evaluating}. However, a common format is provided in \ref{igt} for a line of Old Irish.

\begin{small}
  \begin{exe}
    \ex 
    \gll ní-s-nith anúnas \\
    Neg.3sf.\{N\}\textit{eats} {from above}\\
    \trans `It doesn't eat it (f.) from above.' \\
    \citep{lewis_developing_2010}
    \label{igt}
  \end{exe}
  \end{small}

This IGT format uses three lines for each sentence. The first line is the transcription of the text in the language's orthography. This may be segmented into morphemes, as in the example, but much of the recorded IGT does not do any segmentation. 

The second line provides a gloss for each morpheme. Functional morphemes, or \textbf{grams}, are morphemes such as affixes (such as \textit{-s} in English) and closed-class, functional words (such as \textit{to}) which do not carry their own lexical meaning. Functional morphemes are glossed with their grammatical categories or syntactic function \citep{zhao_automatic_2020}; hence, \textit{ní-} is glossed by \textit{Neg}, as it a negation affix. Lexical morphemes, or \textbf{stems}, are open-class words and stems which carry semantic meaning. They are glossed with their English translation; thus \textit{-nith} is glossed as \textit{eats}. 

The third line of an IGT entry provides a translation in a high-resource language such as English. The words of the translation are not necessarily aligned with specific words in the source language, as languages often express equivalent concepts in differing numbers of words.

The goal of this task is to create a system to predict the gloss line, given the transcription and translation lines. We present a simple baseline system which leaves room for future improvement.

\section{Methods}
\subsection{Data}
The data used in the shared task includes six low-resource languages with various amounts of data, ranging from just 31 sentences in Gitksan to over 39k sentences in Arapaho, and with different features which may include part-of-speech tags and English or Spanish translations. In the open track, sentences are provided with morphological segmentation, while the closed track omits this information. 

\subsection{Model Architecture}
The baseline system utilizes the RoBERTa architecture with the default parameters \citep{liu_roberta_2019}. The task is treated as a token classification task, where each word or morpheme is an input token, and the IGT gloss (or gloss compound) is the label. In the closed track, words are the input tokens; in the open track, morphemes are used. 

A transformer-based architecture is an effective choice for this task, as labelling morphemes often involves disambiguating homophonous morphemes based on context. For example, the English plural morpheme \mbox{\textit{-s}} is spelled the same as the present-tense third-person singular verb morpheme, and the correct label must be determined from the context of the word and sentence. 

We experimented with using a sequence-to-sequence model, but it requires more data to converge, and performs much worse during evaluation, as a single inserted token can cause the following tokens to all be marked incorrect.

Training is done using the AdamW optimizer \citep{loshchilov2017adamw} with beta1 of 0.9, beta2 of 0.999, and epsilon of 1e-8. Models are trained with a learning rate of 2e-5 for 80 epochs, with a batch size of 16 and 0.01 weight decay. Models were trained with an Nvidia V100 GPU and took anywhere from one to twenty hours, depending on language. 

\subsection{Preprocessing}
We tokenize the data using a regular expression to split words or morphemes, omitting punctuation and whitespace. Morphemes are split so that subsequent morphemes after the first one retain their dash, such as the \textit{-s} in \textit{cat-s}, so the model can distinguish word boundaries. The input transcription and translation are encoded using two separate vocabularies and concatenated into a single string with a separator token.

\begin{table*}[!bt]
    \centering
    \def\arraystretch{1.5}
    \begin{adjustbox}{width=\linewidth}
    \begin{tabular}{l l c c c c c c c c c c c}
        \toprule
        \textbf{Language} & \textbf{Track} & \multicolumn{2}{c}{\textbf{Acc. (Morpheme)}} & \multicolumn{2}{c}{\textbf{Acc. (Word)}} & \textbf{BLEU} & \multicolumn{3}{c}{\textbf{Stems}} & \multicolumn{3}{c}{\textbf{Grams}} \\
        & & \multicolumn{1}{c}{Ovr.} & \multicolumn{1}{c}{Avg.} & \multicolumn{1}{c}{Ovr.} & \multicolumn{1}{c}{Avg.} & & P & R & F1 & P & R & F1 \\
        \midrule
        \multirow{2}{*}{Arapaho (arp)} & \multicolumn{1}{l}{Closed} & 43.2 & 51.9 & 70.8 & 70.1 & 41.8 & 50.2  & 48.0 & 49.1 & 63.4 & 33.4 & 43.7 \\
            & \multicolumn{1}{l}{Open} & 91.1 & 91.5 & 85.4 & 85.5 & 79.2 & 91.3 & 89.2 & 90.2 & 91.2 & 94.9 & 93.0 \\\hline
        \multirow{2}{*}{Tsez (ddo)} & \multicolumn{1}{l}{Closed} & 47.5 & 52.9 & 71.8 & 72.1 & 57.8 & 49.7 & 49.2 & 49.4 & 50.7 & 46.1 & 48.3 \\
            & \multicolumn{1}{l}{Open} & 85.0 & 86.0 & 74.2 & 75.8 & 68.6 & 89.3 & 86.6 & 87.9 & 82.2 & 83.6 & 82.9 \\\hline
        \multirow{2}{*}{Gitksan (git)} & \multicolumn{1}{l}{Closed} & 13.6 & 16.3 & 26.5 & 29.1 & 4.5 & 6.7 & 5.8 & 6.2 & 22.2 & 17.6 & 19.7 \\
            & \multicolumn{1}{l}{Open} & 30.0 & 30.2 & 25.0 & 25.7 & 14.2 & 37.8 & 15.0 & 21.5 & 41.8 & 37.8 & 39.7 \\ \hline
        \multirow{2}{*}{Lezgi (lez)} & \multicolumn{1}{l}{Closed} & 48.1 & 49.2 & 56.9 & 55.7 & 52.0 & 54.0 & 51.3 & 52.6 & 53.5 & 40.7 & 46.2 \\
            & \multicolumn{1}{l}{Open} & 50.1 & 52.5 & 32.6 & 39.4 & 42.0 & 61.2 & 48.6 & 54.2 & 50.1 & 53.5 & 51.8 \\ \hline
        \multirow{2}{*}{Nyangbo (nyb)} & \multicolumn{1}{l}{Closed} & 77.1 & 78.2 & 83.9 & 82.4 & 74.2 & 86.2 & 78.7 & 82.3 & 78.6 & 75.3 & 76.9 \\
            & \multicolumn{1}{l}{Open} & 89.2 & 88.5 & 84.7 & 83.6 & 78.4 & 92.5 & 90.5 & 91.5 & 85.9 & 87.6 & 86.8 \\ \hline
        \multirow{2}{*}{Uspanteko (usp)} & \multicolumn{1}{l}{Closed} & 63.1 & 65.5 & 74.0 & 70.3 & 53.8 & 72.0 & 62.7 & 67.0 & 61.3 & 63.7 & 62.4 \\
            & \multicolumn{1}{l}{Open} & 81.3 & 76.2 & 75.9 & 72.0 & 64.9 & 79.4 & 74.0 & 76.6 & 83.7 & 90.4 & 87.0 \\ \bottomrule
    \end{tabular}
    \end{adjustbox}
    \caption{Results of all models}
    \label{tab:results}
\end{table*}

\subsection{Evaluation Metrics}
We evaluate using a variety of metrics. We calculate per-token accuracy for both words and morphemes, including the overall accuracy and average accuracy per sentence. We also calculate F1, precision, and recall across stems and grams. Finally, we calculate BLEU score on the morpheme gloss sequence, which may be favorable to sequence-to-sequence models where tokens might be inserted or deleted.

The exact evaluation script used is available in the task repo.

\section{Results}
The performance metrics for our models for all languages and tracks are listed in \autoref{tab:results}. 

As expected, the models trained with segmentation (open track) outperformed the models without, by up to a 16.4\% improvement in morpheme accuracy. The improvement varied however, with languages such as Lezgi showing minimal improvement–a likely reason is that these languages are less agglutinative and have fewer morphemes per word. 

In the closed track, word-level accuracy tended to be higher than morpheme accuracy. This indicates that the models could learn words with few morphemes more accurately than words with a greater number of constituent morphemes, which were more likely to be out-of-vocabulary. In the open-track, the reverse was true, indicating that even when a model didn't predict an entire word correctly, it often labelled some of the constituent morphemes with the correct gloss. 

\subsection{Future Research}

This baseline model achieves reasonable performance across languages but intentionally leaves much work to be done. 

Critically, the word-level model has no way to make predictions for unseen words, even if the constituent morphemes are known. Learning to infer morpheme segmentation in an unsupervised manner is a difficult task, although some research has had success \citep{palmer_computational_2010, ustun2016unsupervised}. To solve this issue, future models might consider using subword input representations such as characters or byte-pair encoding, although \citet{bostrom_byte_2020} suggests that the latter is not effective at inferring morphology. However, this approach requires a different architecture than token classification, which can present additional difficulties.

The open track models do not utilize any additional information such as part-of-speech tags or external resources, but these could be used effectively to improve model predictions.

All of the models here operate on the orthographic transcription to make predictions. However, morphemes often occur with multiple different allomorphic forms, which may be pronounced similarly but spelled differently. For example, in English the plural morpheme can take the form \mbox{\textit{-s}} or \textit{-es} depending on the stem. A model which incorporates phonetic similarity could learn more efficiently and make better predictions on unseen or ambiguous morphemes.

The models presented in this work encode the translation words using a novel vocabulary. However, since translations are given in a high-resource language, pretrained word embeddings could be used to supplement the translation with contextual information, potentially aiding in glossing.

Finally, the same model architecture was used for every language, despite vast differences in the quantity of available training data. In extremely low-resource settings, it can be beneficial to reduce the size of the architecture (and thereby the hypothesis space); such an approach might show better results on languages like Gitksan \citep{gessler_microbert_2023}.

\section{Conclusion}
In this paper, we presented a simple transformer baseline model for the SIGMORPHON 2023 Shared Task on Interlinear Glossing. Our models treat the task as token classification and achieve varied results across languages and tracks. We present several directions for future research, such as better inferring morphological segmentation and utilizing phonetic similarity between morphemes to make better predictions.

\bibliography{custom}
\bibliographystyle{acl_natbib}

\end{document}